\documentclass{sig-dl2014}

\usepackage{graphicx}
\usepackage{color}
\usepackage{listings}
\usepackage[square,numbers]{natbib}
\usepackage{tabu}
\usepackage[table]{xcolor}
\usepackage{enumitem}
\usepackage{hyphenat}
\begin{document}
\conferenceinfo{Digital Libraries 2014}{London, 8th-12th September 2014}

\title{Extraction of Evolution Descriptions from the Web}

\numberofauthors{2}

\author{
\alignauthor
Helge Holzmann\\
       \affaddr{L3S Research Center}\\
       \affaddr{Appelstr. 9a}\\
       \affaddr{30167 Hanover, Germany}\\
       \email{holzmann@L3S.de}
\alignauthor
Thomas Risse\\
       \affaddr{L3S Research Center}\\
       \affaddr{Appelstr. 9a}\\
       \affaddr{30167 Hanover, Germany}\\
       \email{risse@L3S.de}
}

\crdata{978-1-4799-5569-5/14/\$31.00 \copyright 2014 IEEE}

\maketitle

\begin{abstract}
The evolution of named entities affects exploration and retrieval tasks in digital libraries. An information retrieval system that is aware of name changes can actively support users in finding former occurrences of evolved entities. However, current structured knowledge bases, such as DBpedia or Freebase, do not provide enough information about evolutions, even though the data is available on their resources, like Wikipedia. Our \emph{Evolution Base} prototype will demonstrate how excerpts describing name evolutions can be identified on these websites with a promising precision. The descriptions are classified by means of models that we trained based on a recent analysis of named entity evolutions on Wikipedia.
\end{abstract}





\section{Introduction}

Digital libraries profit from external resources as information source for enriching existing data. This additional knowledge supports users in exploration and information retrieval (IR) tasks. As documents in libraries have been created at different times and eras, current as well as historical knowledge is required, besides information about the evolution. Especially changed names of entities can drastically affect key word search on older texts. IR systems that are aware of current and former names together with the date of the change can actively support users in finding former occurrences of evolved entities. However, while current facts about an entity are available on several knowledge bases, historical knowledge and evolution information are difficult to obtain. We demonstrate a system that provides evolution descriptions, automatically extracted from unstructured websites.

\begin{sloppypar}
An important knowledge source for many applications is Linked Data. With the Semantic Web many knowledge bases consisting of Linked Data have emerged. They provide information of all kinds of entities, which are either manually maintained or automatically extracted from websites. An often used resource for the extraction is Wikipedia. Even though Wikipedia provides quite extensive knowledge, this is largely unstructured and hard to exploit in an automatic way. Therefore, knowledge bases such as DBPedia,
\footnote{http://www.DBpedia.org}
Freebase
\footnote{http://www.Freebase.com}
and Yago
\footnote{http://www.mpi-inf.mpg.de/yago-naga/yago}
focus on available structured information, like info boxes on Wikipedia. While these contain current facts and even some historical data, like former names, they rarely include evolution information, e.g., when and why a name changed. This leads to the lack of evolution data in the above knowledge bases.
\end{sloppypar}

In previous work we found that evolution information are available on Wikipedia, yet hidden in the text and not easy to parse \cite{HolzmannWebSci2014}. We analyzed Wikipedia regarding name evolution of entities. The aim was to understand whether or not Wikipedia can be used as a resource of named entity evolutions. By incorporating lists of name changes, consisting of preceding as well as succeeding names and dates, we identified 62.3\% to be mentioned in the corresponding Wikipedia articles. Moreover, we showed that 79.7\% of them are mentioned in excerpts consisting of less than three sentences. 

Based on our findings we trained classifiers to automatically detect these excerpts in Wikipedia articles and other websites. Our \emph{Evolution Base} demonstrator extracts potential excerpts and classifies them on the fly for a given Wikipedia query or website URL. The excerpts classified as describing an evolution are presented as a timeline of the corresponding entity in chronological order, based on the first year identified within a text. Our demonstration is available on http://evobase.L3S.de/DL2014\_demo.

\section{Approach}

\begin{figure}
\label{fig:screenshot}
\centering
\includegraphics[width=\columnwidth]{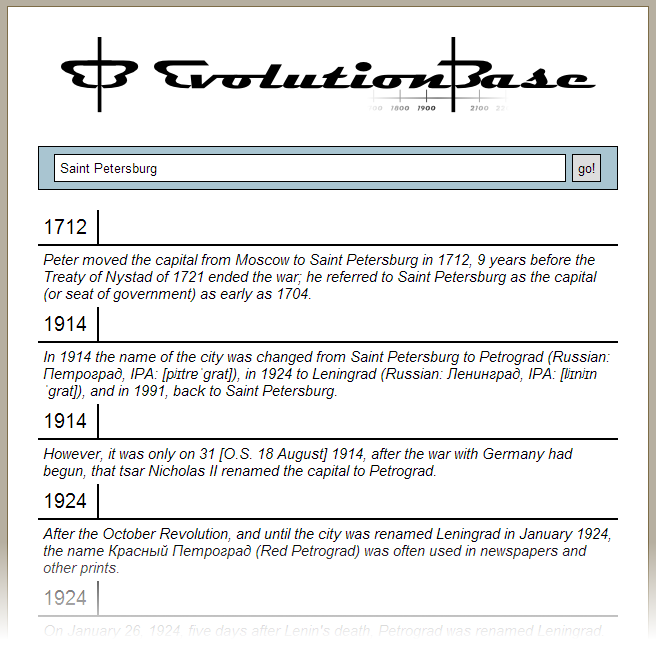}
\vspace{-20pt}
\caption{Screenshot of the Evolution Base demo application showing the results for query \textit{Saint Petersburg}.}
\vspace{-2em}
\end{figure}

The approach we implemented in our demonstration is based on the analysis results presented in \cite{HolzmannWebSci2014}. In this paper we introduced the terminology of sentence distance of a name evolution within an excerpt. It denotes the distance of the first and the last sentence spanned by the components of a name change: preceding name, succeeding name and the change date. As we found more than two-third of all name changes covered within a sentence distance up to 2, we created three training sets for the three excerpt lengths and one consisting of all excerpts, resulting in four classification models.

\subsubsection*{Data}
\label{Datasets}

Every dataset consists of the positive examples that we extracted from Wikipedia and a random sample of negative examples, which are the remaining excerpts of same length from the considered articles. Besides the length we required all excerpts to have certain properties that are characteristic for evolution excerpts. They need to contain at least two names (i.e., preceding and succeeding name) and a year (i.e., the change date). In order to ensure that a change spans the whole excerpt and is not contained in a shorter excerpt within the long one, we required the first and the last sentence of an excerpt to contain one of the three components. For excerpts of distance 0, first and last sentence are the same. In order to identify names, we annotated the words of a sentence using a part-of-speech tagger and considered proper nouns as names. Before extracting the features, all names (N) and years (Y) were replaced by special tokens in order to generalize the excerpts and focus the classification on the mere structure. The same procedure is performed on the fly to select and prepare excerpts from the queried websites or Wikipedia articles.

\subsubsection*{Classification}
\label{sec:classification}

As features we used all ordered word pairs in the excerpts that were identified as describing an evolution. In the sentence ``N evolved and was renamed N in Y'' this would be \textit{N-evolved}, \textit{N-was}, \dots, \textit{was-renamed}, \dots, \textit{in-Y}. In contrast to n-grams, particularly bigrams, the advantage is that we can recognize patterns like ``\textit{N ... renamed N ... Y}'' (pairs: \textit{N-renamed}, \textit{renamed-N}, \textit{renamed-Y}) even though a name and the term \textit{renamed} do not occur next to each other in our example.

The system can work with any classification method, simply by exchanging the model files. For each sentence distance we can specify which classifiers should be used. If more than one classifier is specified, all of them need to classify an excerpt as describing an evolution in order to present it in the result. This dynamic approach enabled us to try different classifiers and combinations. We found the best results achieved by a SVM classifiers that we trained using the SMO algorithm \cite{Platt1998}. A 10-folds cross-validation of the resulted models has shown that they correctly classify between 80.9\% and 93.4\% excerpts on the four datasets. To further improve the results, for each distance we combine the specific classifier with the one that we trained on all distances. Although this lowers the recall by filtering out excerpts that have been correctly classified as describing an evolution by one classifier, more importantly, it filters out the false positive classified excerpts and thus leads to a better precision on identifying evolutions. Since the on-the-fly extraction and classification is a very time-consuming task and took up to a minute in our tests, we cache the results to enable an immediate response on recurring queries. 

The results of the query \emph{Saint Petersburg} are shown in Figure~\ref{fig:screenshot}. The presented excerpts were extracted from the corresponding Wikipedia article and classified as potential evolutions. Such results can be used in a digital library to give users insights into entity evolutions. Other interesting examples are \textit{Mumbai}, \textit{Malawi}, \textit{Edo}, \textit{Muhammad Ali}, \textit{Microsoft Kinect} and more.

\section{Conclusions and Future Work}
\label{sec:conclusions}

Our \emph{Evolution Base} demonstrator shows that excerpts describing name evolutions of entities can be identified with a promising precision. As we solely had geographic name evolution data available for the analysis in \cite{HolzmannWebSci2014}, the classifiers in our demo also perform best on these. However, it also shows that the same patterns can identify name evolutions of different entity types, for instance persons, such as \textit{Muhammad Ali}. The system we developed is a first step towards a knowledge base that covers all kinds of evolutions and would augment existing Linked Data sources with evolution information. In future work we want to investigate different features and classification methods in order to improve the results further. Once we reach a reliably high precision we are able to identify more name changes, which can lead to new excerpts and patterns to extend our training sets. The next step is then to identify the components of a name evolution within the extracted excerpt and provide the results in a structured form. 

\bibliography{bib}{}
\bibliographystyle{unsrtnat}

\end{document}